# SEMANTIC SEGMENTATION FOR URBAN PLANNING MAPS BASED ON U-NET

*Zhiling Guo[1,2], Hiroaki Shengoku[1,2], Guangming Wu[1], Qi Chen[1], Wei Yuan[1], Xiaodan Shi[1], Xiaowei Shao[1*], Yongwei Xu[1], Ryosuke Shibasaki[1]*

[1]Center for Spatial Information Science, University of Tokyo, Kashiwa, Japan
[2]Microbase Inc., Tokyo, Japan

## ABSTRACT

The automatic digitizing of paper maps is a significant and challenging task for both academia and industry. As an important procedure of map digitizing, the semantic segmentation section is mainly relied on manual visual interpretation with low efficiency. In this study, we select urban planning maps as a representative sample and investigate the feasibility of utilizing U-shape fully convolutional based architecture to perform end-to-end map semantic segmentation. The experimental results obtained from the test area in Shibuya district, Tokyo, demonstrate that our proposed method could achieve a very high Jaccard similarity coefficient of 93.63% and an overall accuracy of 99.36%. For implementation on GPGPU and cuDNN, the required processing time for the whole Shibuya district can be less than three minutes. The results indicate the proposed method can serve as a viable tool for urban planning map semantic segmentation task with high accuracy and efficiency.

## 1. INTRODUCTION

The urban planning maps, which provide abundant information of land use, are quite indispensable resource for different fields. However, such kind of maps is often outdated or only nondigital version provided, the digitizing has become a significant task. As a very important part of urban planning maps digitizing, semantic segmentation is mainly relied on manual visual interpretation, which would be very time consuming and inevitably causes many severe problems. The automatic semantic segmentation [1] for urban planning maps with high accuracy and efficiency is still a big challenge.

Except for visual interpretation, previously, researchers mainly focus on conventional image segmentation methodologies [2] for map segmentation, such as graph theory-based [3], clustering-based [4], and classification-based [5] methods. The algorithms like SLIC superpixels [6] and SVM segmentation [7] have shown high feasibility in many specific segmentation tasks [8]. However, due to the complexity and variety of map background, adequate deep and abstract patterns cannot be well represented by the traditional graphic and hand-crafted textural features, which easily leads to poor generalization.

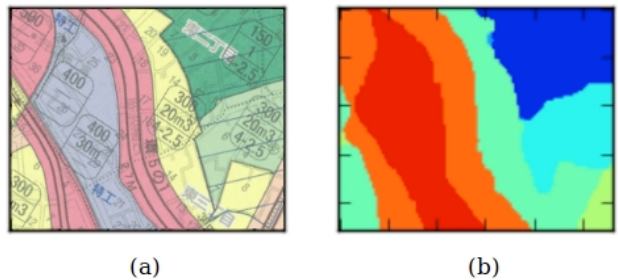

**Fig. 1**. Semantic segmentation for urban planning maps based on U-Net. (a) Urban planning map. (b) Semantic segmentation result by optimized U-Net.

With the rapid development of machine learning techniques in recent years, a considerable number of deep learning [9] based segmentation methods have been applied in different fields with tremendous effect [10-12]. To the best of our knowledge, there is no existing empirical research addresses the semantic segmentation of urban planning maps via deep learning. In this study, we investigate the feasibility of utilizing U-shape fully convolutional (U-Net) [13] architecture to perform end-to-end and pixel-to-pixel map semantic segmentation for urban planning maps. As shown in Fig. 2, an automatic map digitizing platform is elaborately developed to provide rapid, accurate, and efficient semantic segmentation results for urban planning maps.

The main contributions of this study are listed as follows. First, this is the first research to investigate the feasibility of automatic map digitizing based on deep learning framework. Second, we combined deep learning with other image processing methodologies, and obtained an optimized U-Net model that can adapt to urban planning maps semantic segmentation based on rigorous experiments. Third, the proposed method is proven to be an efficient tool for urban planning maps semantic segmentation and achieved very high Jaccard similarity coefficient results in Shibuya district,

Tokyo. Last but not least, except for urban planning maps, the semantic segmentation system could be potentially utilized in other map semantic segmentation tasks.

## 2. DATA

To evaluate the feasibility and robustness of the proposed method, here, we deliberately selected a complicated and typical area in Tokyo: Shibuya district, which contains eleven different types of land usage areas. The diversity and complexity of the maps also make the segmentation task difficult. This, in turn, warrants that the segmentation model incorporate all these conditions. The RGB urban planning maps of Shibuya district and corresponding ground truth images are prepared by scanning paper version old maps and manually drawn beforehand annotations respectively. The ground truth contained accurate information of the land categories and was chiefly used for sampling and result detection. We divided the whole dataset into three parts to implement training, cross validation, and testing.

## 3. METHODOLOGY

In this study, a novel method for urban planning maps semantic segmentation is proposed based on U-Net and a few of image processing methodologies. Fig. 2 shows the scheme of the study. The urban planning maps images of study area undergo a process of data preprocessing to generate training and testing data. Our proposed U-Net model is trained and cross validated using training data. The trained model with proper hyperparameters that pass the cross validation, will be chosen to make prediction on testing data and then be evaluated by two commonly used performance indexes of Jaccard similarity coefficient and overall accuracy. Finally, a few of post processing methods were implemented to enhance semantic segmentation results.

### 3.1. Data pre-processing

Artificially enlarge the dataset using label-preserving transformations is the most common method to reduce over-fitting. Here, we employed data augmentation to increase the diversity of training dataset. Considering the diversity of urban planning maps, we enrich the category by rotating or stretching the training samples; in that case, many new patches of maps in different structure will be generated with accurate ground truth. The preceding data augmentation is randomly implemented when collecting training samples, and 10% new samples are generated.

### 3.2. Model

*3.2.1. U-shape full convolutional*
Compared to traditional image segmentation methods, the U-Net model is more robust and can yield better performance in image segmentation owing to its capability to apply end-to-end and pixel-to-pixel mapping by mining deep and representative information from low-level inputs. Rather than the fully convolutional networks (FCN) [14], the U-Net model strongly use data augmentation via adopting skip connections and top-down/bottom-up design architecture, which combines lower layers and higher layers to generate final output, that results in better performance. Like convolutional neural networks (CNN) [15], the U-Net model also performs the steps of convolution, non-linear activation and pooling.

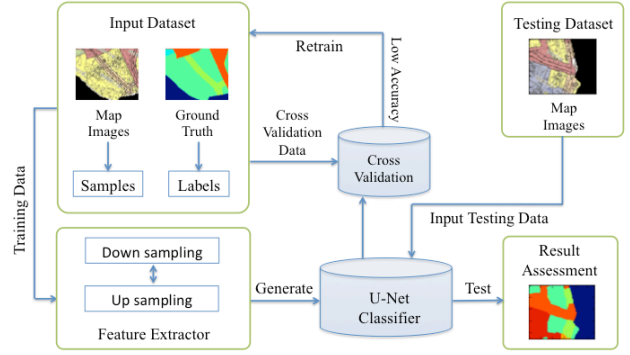

**Fig. 2**. Workflow of map semantic segmentation platform

In this experiment, the input dataset $x$ in size $h \times w \times c$ refers to multichannel urban planning maps images, where each dimension represents the height, width, and number of channels.

The convolutional kernel in middle can generate the output $z$ as follows, where $b$ and $\theta$ refer to the bias and filter kernels weight respectively.

$$z_{h',w',k} = b + \sum_{i=1}^{r1} \sum_{j=1}^{r2} \sum_{d=1}^{c} \theta_{ijdk} \times x_{h'+i,w'+j,c'+d} \qquad (1)$$

The unsaturated activation method Rectifier (ReLU) [16] and max-pooling [17] are applied in U-Net to generate the hypothesis and enlarged the receptive field respectively as follows:

$$y_{h',w',k} = \max(0, z_{h',w',k}) \qquad (2)$$

$$y_{h',w',k} = \max_{1 < i < \tilde{h}, i < j < \tilde{w}} x_{h'+i,w'+j,c'} \qquad (3)$$

Finally, the segmentation result can be generated using softmax function [18]:

$$y_{i,j,k} = \frac{\exp(x_{i,j,k})}{\sum_{d=1}^{c} \exp(x_{ijd})} \qquad (4)$$

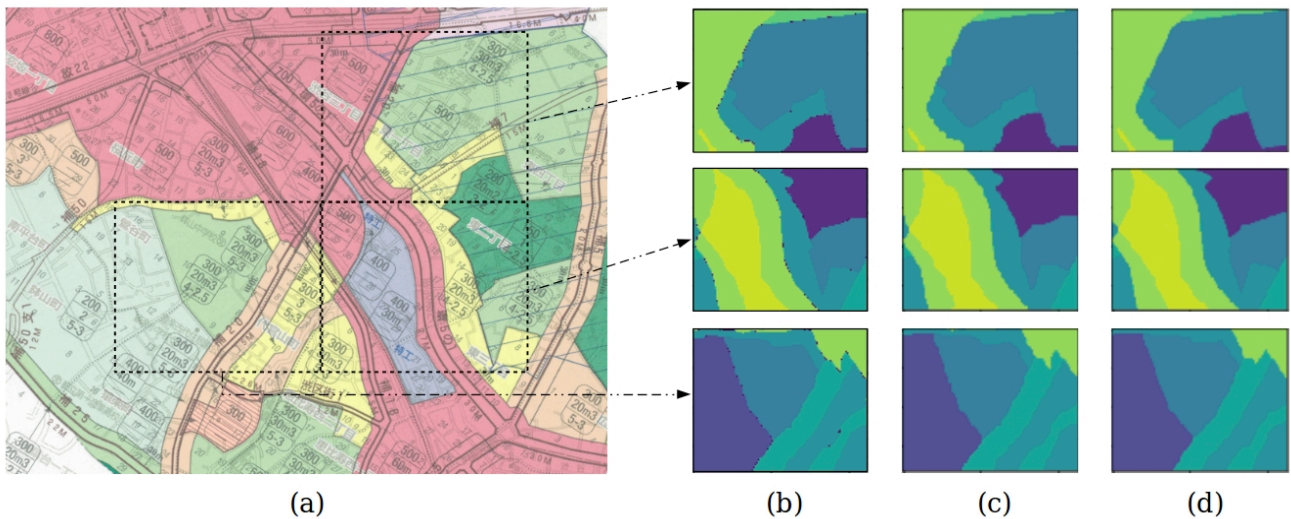

**Fig. 3**. Semantic segmentation results for urban planning maps. (a) Urban planning map of Shibuya district, Tokyo. (b) Semantic segmentation results by optimized U-Net. (c) Denoised results. (d) Ground truth images.

The tricks such as batch normalization (BN) [19] and ReLU are heavily applied after every convolution layer to accelerate deep network training and avoid vanishing gradient problem. With these layers trained by stochastic gradient descent (SGD) [20] and back propagation (BP) algorithms, the proposed U-Net model can learn a pattern from input RGB map images to generate equal size segmentation image.

*3.2.2. Model optimization*
The original U-Net with strong feature extraction capability is typically used for biomedical image segmentation in specific image size. Considering the characteristics of the input urban planning maps and our segmentation targets, we optimized the structure of original U-Net based on analyzing the learning curves and bias/variance of the training/cross validation results. Intuitively, the optimized U-Net owns fewer filters in middle layers, while more BN layers are applied. Also, we resized the input patch size and modified the output tensor into eleven categories to match the segmentation requirement.

**3.3. Post Processing**

The semantic segmentation results directly generated by U-Net inevitably contain the salt-and-pepper noise especially near the boundary between different land usage areas. Here, we denoise and smooth the semantic segmentation results by implementing the kernel based morphological methods.

## 4. RESULTS AND DICUSSIONS

Fig. 3 shows part of the semantic segmentation results for Shibuya district in Tokyo. According to the principle, the urban planning map images are input into a trained U-Net model for feature extraction and segmentation. The rough semantic segmentation results, which could significantly separate different land usage regions, can be generated as shown in Fig. 3 (b). Thereafter, by applying morphological methods, the salt-and-pepper noise near the boundary, which heavily affects the results can be easily removed. Compared with the ground truth in Fig. 3 (d), the final denoised experimental results (Fig. 3 (c)) of the testing area at Shibuya district in Tokyo indicate that our proposed method can achieve a Jaccard similarity coefficient of 93.63% and a very high overall accuracy of 99.36% for semantic segmentation. Moreover, The total processing time for both training and testing procedure could be less than three minutes by applying GPGPU and cuDNN. The performance indicates the proposed method can be a viable tool for urban planning map semantic segmentation task with high accuracy and efficiency.

It should be noted that a further analysis and comparison of the model's feasibility in all cases and the regions that are not included in this study is every difficult, because they differ in terms of data acquisition methods, reference datasets, and class definitions.

## 5. CONCLUSIONS

In this research, we presented a novel U-Net based method to conduct semantic segmentation for urban planning maps. As far as we know, this is the first study to implement the deep learning framework for urban planning map digitizing. The high efficiency and accuracy of the results obtained from Shibuya district, Tokyo, demonstrate the proposed method can be a feasible map digitizing and segmentation tool for urban planning maps. Moreover, we believe that the proposed method could be potentially implemented in semantic segmentation tasks for other regions and map types, not merely urban planning maps. For future works, the techniques such as deep learning based optical character recognition (OCR) [21] will be applied and combined with proposed segmentation method to achieve fully-automatic map digitizing.

## 6. ACKNOWLEDGEMENTS

This work was supported by The Japan Society for the Promotion of Science (JSPS) Grant (No. 16k18162), China Postdoctoral Science Foundation with Project Number 2016M590730, and the National Natural Science Foundation of China with Project Number 41601506. The ground truth data was provided by Mircobase Inc.